\documentclass[10pt,twocolumn,letterpaper]{article}

\usepackage[pagenumbers]{cvpr} 

\definecolor{cvprblue}{rgb}{0.21,0.49,0.74}
\usepackage[pagebackref,breaklinks,colorlinks,allcolors=cvprblue]{hyperref}

\usepackage{algorithm}
\usepackage{algpseudocode}
\usepackage{array}
\usepackage{subcaption}
\usepackage{multirow}

\def\paperID{11562} 
\def\confName{CVPR}
\def\confYear{2025}

\title{RAD: Region-Aware Diffusion Models for Image Inpainting}

\author{
Sora Kim\\
Hanyang University\\
Ansan, Republic of Korea\\
{srk1995@hanyang.ac.kr}
\and
Sungho Suh\\
DFKI and RPTU Kaiserslautern-Landau\\
Kaiserslautern, Germany\\
{sungho.suh@dfki.de}
\and
Minsik Lee\\
Hanyang University ERICA\\
Ansan, Republic of Korea\\
{mleepaper@hanyang.ac.kr}
}

\begin{document}

\twocolumn[{%
\renewcommand\twocolumn[1][]{#1}%
\maketitle
\vspace{-5mm}
\centering
\includegraphics[width=\linewidth]{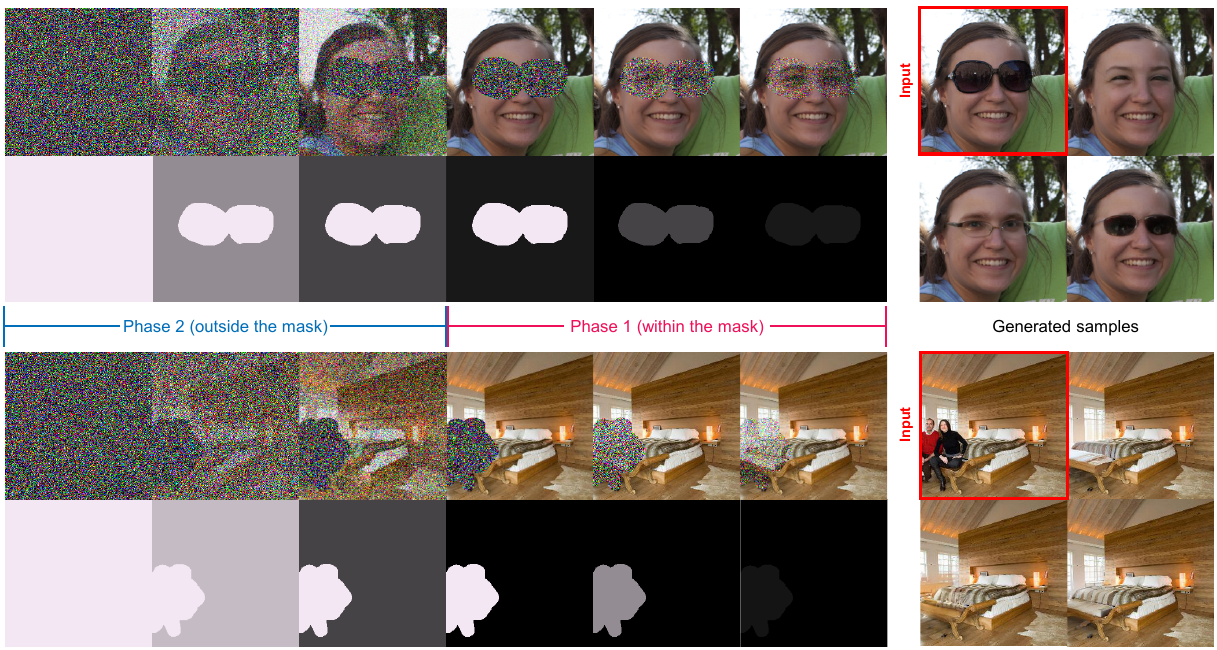} \label{fig:overview}
\vspace{-5mm}
\captionof{figure}{Region-aware diffusion models (RAD) in action.}
\vspace{4mm}
\label{fig:teaser}
}]


\begin{abstract}
Diffusion models have achieved remarkable success in image generation, with applications broadening across various domains.
Inpainting is one such application that can benefit significantly from diffusion models. 
Existing methods either hijack the reverse process of a pretrained diffusion model or cast the problem into a larger framework, \ie, conditioned generation. However, these approaches often require nested loops in the generation process or additional components for conditioning. 
In this paper, we present region-aware diffusion models (RAD) for inpainting with a simple yet effective reformulation of the vanilla diffusion models.
RAD utilizes a different noise schedule for each pixel, which allows local regions to be generated asynchronously while considering the global image context. 
A plain reverse process requires no additional components, enabling RAD to achieve inference time up to 100 times faster than the state-of-the-art approaches. Moreover, we employ low-rank adaptation (LoRA) to fine-tune RAD based on other pretrained diffusion models, reducing computational burdens in training as well.
Experiments demonstrated that RAD provides state-of-the-art results both qualitatively and quantitatively, on the FFHQ, LSUN Bedroom, and ImageNet datasets.
\end{abstract}
\vspace{-5mm}    
\section{Introduction}
\label{sec:intro}
Over the past decade, deep generative models \cite{kingma2013auto, goodfellow2014generative, sohl2015deep, ho2020denoising} have made significant advances in generative learning. Especially, generative adversarial networks (GANs) \cite{goodfellow2014generative} and diffusion models \cite{sohl2015deep, ho2020denoising} represent seminal breakthroughs that have changed the paradigm of unsupervised image synthesis. Recently, diffusion models have attracted considerable attention due to their outstanding performance across various applications, such as text-to-image synthesis \cite{rombach2022high, saharia2022photorealistic, nichol2021glide}, image editing \cite{meng2021sdedit, chen2024don}, video frame generation \cite{yang2023diffusion}, and text-to-3d generation \cite{poole2022dreamfusion, yi2023gaussiandreamer}.

Denoising diffusion probabilistic models (DDPM) \cite{sohl2015deep, ho2020denoising}, a cornerstone in diffusion models, approximate the distribution of real images by learning to reverse a pre-defined diffusion process in which real images gradually become pure Gaussian noise. Based on the learned reverse process, synthetic images can be generated by iteratively denoising arbitrary Gaussian noise. In this process, each reverse step is represented as a Gaussian transition and is modeled with deep networks like U-Net \cite{ronneberger2015u}. Even though DDPM had a compelling framework, it initially produced lower fidelity results than GANs.
Various studies \cite{nichol2021improved, xiao2022DDGAN, phung2023wavediff} followed to improve the performance of diffusion models, and Dhariwal \etal \cite{dhariwal2021diffusion} proposed the first diffusion model to outperform GANs.
Nowadays, diffusion models have become \emph{de facto} standard for image generation, and numerous applications have been inspired by their well-established theory and outstanding performance.

Image inpainting, a problem to fill in missing areas of an image, is such an example that can benefit largely from a powerful generative model. With the notable advancements of deep generative models, many attempts have been made to solve image inpainting based on GANs \cite{liu2018image, yu2019free, suvorov2022resolution, wu2024syformer} and diffusion models \cite{lugmayr2022repaint, couairon2023diffedit}. 
Diffusion-based methods have proven effective for various image inpainting or editing tasks, and some approaches utilize conditioned generation techniques based on structural information \cite{liu2024structure} or text with mask \cite{yang2023paint, xie2023smartbrush, zhuang2023task, zhao2024udifftext}. These methods have advantages in providing more precise control over the result, but they generally require additional modules to process conditions, which add more complexities and computational burdens. 

Another line of research focuses on manipulating the generation processes in existing diffusion models. These approaches \cite{meng2021sdedit, chung2022improving, hertz2022prompt, couairon2023diffedit, avrahami2023blended, corneanu2024latentpaint, lugmayr2022repaint, kawar2022denoising, wang2023ddnm, cao2024deep} `hijack' the reverse process of a pretrained diffusion model
and devise elaborate procedures for inpainting or editing.
These methods do not require any additional training, however, since the plain reverse process is not usually designed for localized generation or editing, the procedures tend to get complicated, \eg, requiring repeated re-evaluation of reverse steps, etc., resulting in significantly extended inference time.

In this paper, we propose region-aware diffusion models (RAD), a simple reformulation of the vanilla diffusion models to overcome the aforementioned issues in diffusion-based image inpainting. Unlike conventional diffusion models, which apply noise uniformly across all pixels at each forward step, RAD assigns a different noise schedule to each pixel, enabling some areas to be completely denoised while others retain noise. This spatially variant noise scheduling naturally emulates inpainting by adding noise only to the inpainting region and performing the reverse process.
This idea is quite simple, and only minimal changes in the network structures, \ie, `reshaping' some fully connected (FC) layers into $1\times1$ convolutions, suffice to achieve state-of-the-art (SoTA) performance, demonstrating that existing structures are readily capable of inpainting once the right setting is provided.
Unlike the existing methods tempering the noise in diffusion models, such as RePaint \cite{lugmayr2022repaint} and MCG \cite{chung2022improving}, RAD inherently considers the asynchronous generation of pixels, achieving orders of magnitude improvement in generation speed while maintaining high performance.

That being said, several issues need to be carefully addressed in RAD. The pixel-wise noise schedules must be designed to represent realistic inpainting patterns and must be somehow informed to the network for effective noise inference. We deal with these issues with simple, novel ideas, \ie, Perlin noise-based schedule generation and spatial noise embedding, respectively. One limitation of the proposed method is that it requires fresh training for the altered diffusion framework; however, we overcome this by employing low-rank adaptation (LoRA) \cite{hu2021lora} on a pretrained model, greatly reducing computational requirements.

We conducted experiments on FFHQ \cite{karras2019style}, LSUN Bedroom \cite{yu2015lsun}, and ImageNet \cite{deng2009imagenet}, comparing RAD with other SoTA inpainting methods. RAD achieves up to 100 times faster inference time than other SoTA diffusion-based methods and achieves the best FID and LPIPS scores in most cases. In addition, an ablation study shows that the proposed components of RAD, such as the spatially variant noise schedules and the spatial noise embedding technique, are vital for the success of RAD.
The contributions of this paper are summarized as follows:
\begin{itemize}
\item A novel reformulation of diffusion models is proposed based on spatially variant noise schedules, allowing asynchronous generation of pixels for inpainting.
\item Pseudo-realistic noise schedules are presented based on Perlin noise for efficient training.
\item A spatial noise embedding technique is introduced to provide rich spatial information to the denoiser networks.
\item Along with the SoTA performance and the exceptional improvement in generation speed, LoRA-based training on pretrained diffusion models is also utilized to reduce the training burdens.
\end{itemize}
\section{Related Works}
\label{sec:related_works}
\paragraph{Diffusion model.} 
DDPMs \cite{sohl2015deep, ho2020denoising} have introduced a novel approach in image generation by employing an iterative denoising process that progressively refines random noise into high-quality images. Unfortunately, DDPMs showed lower image fidelity compared to GANs. To enhance fidelity, various studies \cite{nichol2021improved, xiao2022DDGAN, phung2023wavediff} have focused on generating high-quality images using diverse datasets such as ImageNet \cite{deng2009imagenet}, FFHQ \cite{karras2019style}, and LSUN \cite{yu2015lsun}. Dhariwal and Nichol \cite{nichol2021improved} showed that achieving high log-likelihood on datasets with high diversity, like ImageNet \cite{deng2009imagenet}, is possible through a hybrid objective. This hybrid objective facilitates learning the variances of the reverse Gaussian transitions that were fixed in DDPM \cite{ho2020denoising}. Meanwhile, Dhariwal \etal \cite{dhariwal2021diffusion} introduced ADMs that use auxiliary classifiers to classify the noisy images generated during the reverse process. This simple class-conditioned generation method, known as \emph{classifier guidance}, mainly enhances the fidelity of generated images by applying strong class conditioning. 

\vspace{-4mm}
\paragraph{Diffusion-based inpainting.}
Recently, many conditional image inpainting/editing methods have been proposed based on diffusion models. Liu \etal \cite{liu2024structure} used structural information, such as grayscale images or edge maps, to guide an inpainting process. Several studies \cite{yang2023paint, xie2023smartbrush, zhuang2023task, zhao2024udifftext} have incorporated text and mask conditions for image inpainting. 
These methods typically require additional modules to perform local editing based on the specified conditions, which introduces additional complexity and computational load.

Other approaches leveraged pretrained diffusion models for image inpainting/editing by manipulating their generation processes. Some methods \cite{meng2021sdedit, chung2022improving} manipulated the reverse SDE procedure of a pretrained score-based model \cite{song2021scorebased}. 
On the other hand, several works \cite{lugmayr2022repaint, kawar2022denoising, wang2023ddnm, cao2024deep} utilize ADM \cite{dhariwal2021diffusion}. Notably, Lugmayr \etal \cite{lugmayr2022repaint} proposed RePaint, a method utilizing resampling steps to harmonize mask and non-mask regions. Other studies \cite{hertz2022prompt, couairon2023diffedit, avrahami2023blended, corneanu2024latentpaint} used the stable diffusion model \cite{rombach2022high}. Couairon \etal \cite{couairon2023diffedit} proposed DiffEdit, which employs DDIM inversion for image editing by generating masks based on text prompts to preserve backgrounds. 

Existing methods often require additional modules or extended reverse processes, increasing complexity and inference time. 
In contrast, RAD utilizes spatially variant noise schedules, inherently allowing detailed generation in specific areas without any additional component or loss. 
SmartBrush \cite{xie2023smartbrush}, although the problem setting differs from ours, is another method that adds noise only in the inpainting regions. SmartBrush, however, adds several additional modules to a diffusion model to learn this type of noise, unlike ours where the basic framework inherently supports this.
\begin{figure}
    \includegraphics[width=\linewidth]{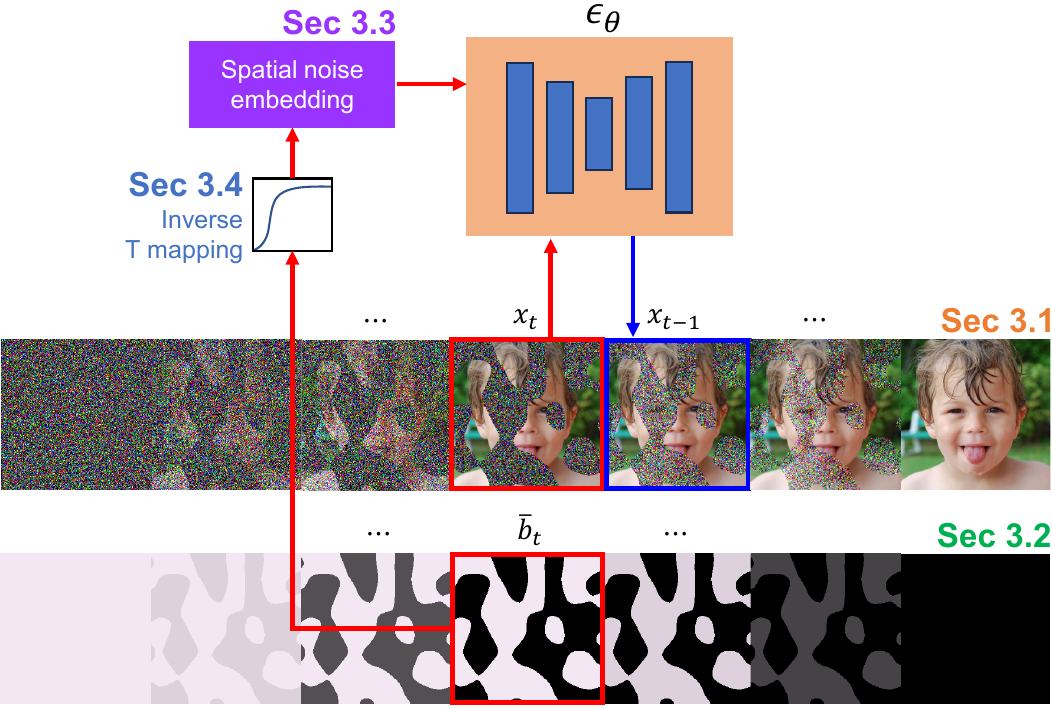} 
    \captionof{figure}{An overview of the proposed method. RAD consists of four components: (1) the forward and reverse processes based on pixel-wise noise (\Cref{sec:RAD}); (2) spatially variant noise schedules (\Cref{sec:noise_schedule}); (3) spatial noise embedding  (\Cref{sec:noise_encoding}); and (4) the inverse-mapping of $\bar{b}$ (\Cref{sec:practical}).}
    \label{fig:proposed_method_overview}
\end{figure}

\section{Region-Aware Diffusion Models}
\label{sec:proposed_method}

The core of the region-aware diffusion models (RADs) is redefining diffusion models so that each pixel has a different noise schedule, emulating inpainting scenarios. RADs define spatially variant noise schedules based on given masks (\cref{sec:noise_schedule}), which are utilized to establish both the forward and reverse processes based on pixel-wise noise (\cref{sec:RAD}). This involves an element-wise reformulation of DDPM, of which many parts of the derivation naturally follow, however, there are some important points to consider (\cref{sec:noise_encoding,sec:practical}). This section will focus particularly on these critical points.
The overview of RAD is shown in \Cref{fig:proposed_method_overview}.

\subsection{Diffusion Models with Spatially Variant Noise} \label{sec:RAD}

Here, we explain the basic framework of RAD, along with a brief explanation of DDPM \cite{ho2020denoising}.
The goal of a diffusion model is to learn to mimic a data distribution $q(x_0)$, where $x_0$ denotes a `clean' image without any noise, based on some model $p_\theta(x_0)$ in an unsupervised manner. This is accomplished by learning an iterative denoising process that reverts a Brownian motion where $x_0$ gradually becomes pure Gaussian. This framework can be more tractable than directly learning $q(x_0)$ because each individual task is to slightly denoise the noisy image at various noise stages.

Given a noise schedule, a sequence $x_1, x_2, \dots, x_T$ can be generated with some $T$ where the image gradually becomes noisier. The \emph{forward} process governing this is defined as a Markov process with Gaussian transitions:
\begin{equation}\label{eq:forward basic}
        q\left(x_{1:T}|x_0\right) = \prod_{t\ge1} q\left(x_{t}|x_{t-1}\right).
\end{equation}

In DDPM, $q\left(x_{t}|x_{t-1}\right) = \prod_i q\left(x_{t, i}|x_{t-1, i}\right)$, $i$ denoting the element index, is assumed to be \iid for the elements of $x_t$.
At this point, RAD takes a different turn from DDPM, assuming the elements have different noise intensities:
\begin{equation}\label{eq:forward element}
         q\left(x_{t, i}|x_{t-1, i}\right) = N\left(\sqrt{1 - b_{t, i}}x_{t-1, i}, \, b_{t, i} \, \right)
\end{equation}
where $b_{t, i}$ is the variance of the $i$-th element of $x_t$ given $x_{t-1}$, \ie, the covariance matrix of $q\left(x_t|x_{t-1}\right)$ is $\text{diag}(b_t)$.
Similarly, the marginal distribution of $x_{t,i}$ can be given as
\begin{equation}\label{eq:forward element with a}
         q\left(x_{t, i}|x_{0, i}\right) = N\left(\sqrt{\bar{a}_{t, i}}x_{0, i}, 1 - \bar{a}_{t, i}\right)
\end{equation}
with $a_{t, i} \triangleq 1 - b_{t, i}$, $\bar{a}_{t, i} \triangleq \prod_{s=1}^t a_{s, i}$, and $\bar{b}_{t, i} \triangleq 1 - \bar{a}_{t, i}$. This spatially variant noise assumption allows asynchronous generation of regions within the image, \ie, different regions can be subjected to different noise intensities so that the individual regions can be generated at different speeds, not affecting already generated ones. This alternate formulation, despite its simplicity, has proven to be quite effective in our experiments.

Even though the forward process is quite simply defined, the \emph{reverse} process, \ie, the denoising process, has no closed-form solution. The essential reason is that the data distribution $q(x_0)$ is not simple, and hence, the Bayes equation cannot be solved easily. Accordingly, we need to learn a model $p_\theta(x_0)$ that approximates the reverse process. In diffusion models, $p_\theta(x_0)$ is also formulated as a Markov process, starting from $x_T$, reversing the timesteps:
\begin{equation}\label{eq:reverse process}
        p_\theta\!\left(x_{0:T}\right) = p\!\left(x_T\right) \prod_{t\ge1} p_\theta\!\left(x_{t-1}|x_t\right), 
\end{equation}
where $p\!\left(x_T\right) = N\!\left(0, I\right)$ is pure Gaussian noise. The reverse transition is also modeled as Gaussian, \ie, $p_\theta\!\left(x_{t-1}|x_t\right) = N\!\left(\mu_{\theta,t}\!\left(x_t\right), \text{diag}(s_t)\right)$, however, the mean $\mu_{\theta,t}$ is a learnable function of $x_t$. 
In practice, the true $\mu_{t}$ of the denoising process can be expressed in terms of $x_t$ and the accumulated noise $\epsilon_t$ (in $x_t$), $\epsilon_\theta(x_t,t)$ is learned to predict $\epsilon_t$ instead:
\begin{equation}\label{eq:mu}
        \mu_{\theta,t,i}\!\left(x_t\right) = \frac{1}{\sqrt{a_{t,i}}} \left(x_{t,i} - \frac{b_{t,i}}{\sqrt{1-\bar{a}_{t,i}}} \epsilon_{\theta,i}(x_t,t) \right).
\end{equation}
Again, this is an element-wise version that can include RAD, while all $a_{t,i}$, $b_{t,i}$, and $\bar{a}_{t,i}$ terms are identical for $i$ in DDPM, and the same goes for $s_t$ as well. In RAD, a division-by-zero can happen in the above equation, \ie, $\bar{a}_{t,i}=1$, if no noise has been added to the $i$-th pixel until $t$. In this case, a separate derivation gives $\mu_{\theta,t,i} = x_{t,i}$.

In the above, although the forward process in \cref{eq:forward basic} has no spatial dependence, $p_\theta\!\left(x_{t-1}|x_t\right)$ surely has. This is because $q(x_0)$ also likely has strong spatial dependence. Accordingly, $p_\theta$ must take the global context into account in the denoising process. This is why $\mu_{\theta,t,i}\!\left(x_t\right)$ and $\epsilon_{\theta,i}(x_t,t)$ in (\ref{eq:mu}), even though they indicate specific (the $i$-th) elements of $\mu_{\theta,t}$ and $\epsilon_{\theta}$, respectively, take the entire $x_t$ as an input.

Given the above forward and reverse processes, a variational loss can be defined as
\begin{equation}\label{eq:reduced_VLB}
\resizebox{0.9\columnwidth}{!}{
    $\mathbb{E}_q\bigg[\sum_{t>1} D_{KL}\!\left(q\!\left(x_{t-1}|x_t, x_0\right) \| \ p_\theta\!\left(x_{t-1}|x_t\right)\right) - \log p_\theta(x_0|x_1) \bigg]$
}
\end{equation}
This basically trains $p_\theta\!\left(x_{t-1}|x_t\right)$ to match the target $q\!\left(x_{t-1}|x_t, x_0\right)$, except for $t=1$.
In \cite{ho2020denoising}, a simpler loss has been proposed as
\begin{equation}\label{eq:simplified_loss}
    \begin{split}
    &L = \sum_{t\ge1} \mathbb{E}_{q}\!\left[\left\| \epsilon_t - \epsilon_\theta(x_t,t) \right\|^2\right],
    \end{split}
\end{equation}
which directly trains $\epsilon_\theta$ to predict $\epsilon_t$. This has shown to be effective, and in many later works \cite{nichol2021improved, dhariwal2021diffusion}, both (\ref{eq:reduced_VLB}) and (\ref{eq:simplified_loss}) have been frequently used. In RAD, the calculation of (\ref{eq:reduced_VLB}) must be done with the element-wise versions of $q$ and $p_\theta$. After training, an image can be generated by performing a reverse process based on the learned $p_\theta$ from a randomly generated Gaussian noise.

As can be seen above, the basic framework of RAD is quite simple, \ie, an element-wise reformulation of DDPM. However, this is an easier part of RAD. For this framework to actually work, there are several important issues to consider: (i) What is an appropriate choice for the spatially variant noise ($b_t$)? (ii) How can $\epsilon_{\theta}$ successfully learn from the altered problem setting?
(iii) How can we reduce additional efforts in training this alternate formulation?
For the rest of the section, we will elaborate on the above points.

\subsection{Generating Noise Schedules} \label{sec:noise_schedule}
Seeing RAD, one can quickly notice that the newly introduced flexibility requires careful attention. Unlike DDPM where a fixed noise schedule is applied to all pixels, those are set differently in RAD to allow asynchronous generation of different regions. These schedules must be somehow designed, and they must include various patterns during training so that diverse inpainting scenarios can be handled.

To encompass various noise shapes, we generate $b_t$ randomly during training. Now, the noise schedules also form a distribution, and the loss function (\ref{eq:reduced_VLB}) must include this in the expectation. In other words, we now have $q(x_{0:T}|b_{1:T})$ where $b_{1:T}$ follows some $q(b_{1:T})$. There are infinite choices for $q(b_{1:T})$, which becomes important since it can affect training. A na\"ive approach, such as selecting a random schedule for each pixel independently, can be problematic because the resulting noise may not have any distinctive spatial pattern, which is inconsistent with the actual inpainting scenarios. Indeed, this was not very successful in our empirical experience, meaning that having too random a noise pattern can be detrimental to the success of RAD.

Considering the above point, we limit the possible shapes of $b_t$ strictly to the inpainting scenarios. Specifically, we divide the entire diffusion process into two phases, where noise is filled in only for the pixels in a given inpainting mask in Phase 1 and the rest in Phase 2. This adequately represents an inpainting process, where only a part of an image is generated (Phase 1) while the other parts are already present (phase 2). The order of these phases is set in reverse order because the actual generation is performed in the reverse process. After training, only Phase 1 is utilized in inpainting because this suffices to generate the mask region. In fact, we may only utilize Phase 1 during training as well, but using both phases was better in our experience.

To mimic the noise-filling process of DDPM in each phase, we use the following strategy: Let $T_1$ and $T_2$ be the numbers of timesteps for Phases 1 and 2, respectively, \ie, $T_1+T_2=T$. During each phase, each pixel in or outside the mask is filled in by a scalar noise with variance $\beta_t$ ($1 \leq t \leq T_1$) or $\beta_t'$ ($1 \leq t \leq T_2$), respectively, with $0 < \beta_t, \beta_t' < 1$. In practice, we use simple linear schedules for $\beta_t$ and $\beta_t'$ as in DDPM. A caveat here is that all pixels must have the same accumulated noise levels after finishing the two phases, which can be satisfied easily by normalizing $\beta_t$ and $\beta_t'$ as explained in the supplementary material.
Gathering the above pixel-wise noise schedules, we can form $b_t$.

\begin{figure}
    \centering
    \includegraphics[width=0.8\linewidth]{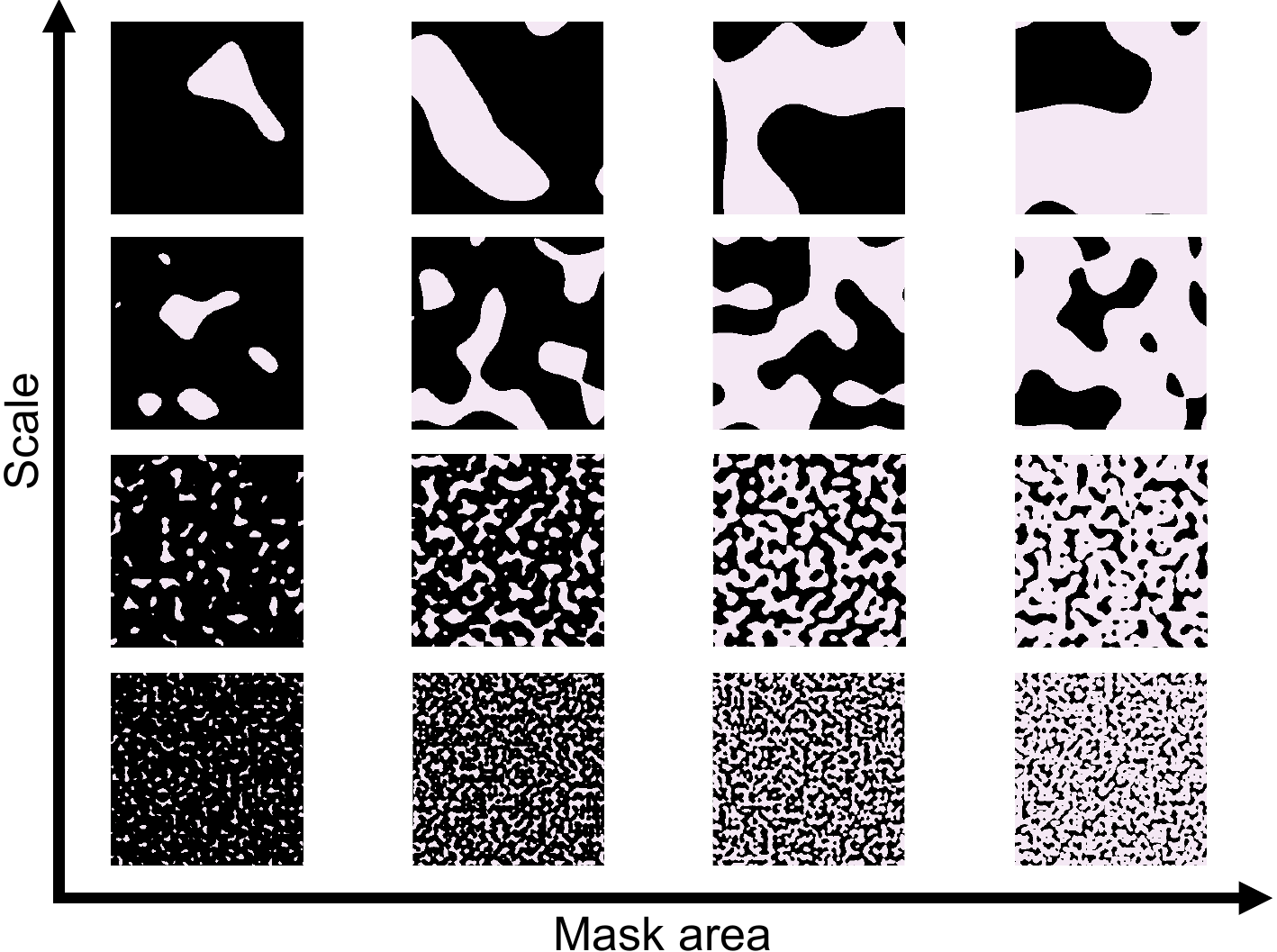}
    \caption{Examples of inpainting masks based on Perlin noise.}
    \label{fig:perlin}
\end{figure}

In the above strategy, the quality of inpainting masks during training is crucial. Unlike test conditions,
numerous masks must be automatically supplied during training. Accessing the true mask distribution is not viable, so we need to find some sort of surrogate that includes diverse natural patterns mimicking real-world inpainting tasks.
To this end, we propose to use Perlin noise \cite{perlin1985image}. Perlin noise, known for its smooth and naturalistic patterns, allows us to create diverse and realistic masks.
To generate binary masks, we utilize black-and-white Perlin noise, which can be obtained by simple thresholding.
We sample Perlin noise with various spatial scales by uniformly sampling the scale parameter so that the generated patterns have both finer and coarser structures. We also sample the black-and-white conversion threshold to control the overall area of inpainting.
\Cref{fig:perlin} shows examples of generated masks. 
This surrogate distribution is quite effective, providing SoTA performance.

An interesting fact is that, even though there is a sharp separation between the mask and non-mask regions in the above strategy, the inpainting results do not exhibit any noticeable boundary effects. We have also tried blurring the boundaries of the masks, with no meaningful improvement in performance. This suggests that RAD is inherently capable of generating content adaptively to existing regions.

\subsection{Spatial Noise Embedding} \label{sec:noise_encoding}
In diffusion models, a deep network $\epsilon_\theta(x_t,t)$ is trained to estimate the accumulated noise $\epsilon_t$ in $x_t$. RAD does not have any structural difference in this regard, \ie, it shares the same input and output for $\epsilon_\theta$.\footnote{In fact, the difference resides in $\mu_{\theta,t}$, where $a_t$, $b_t$, and $\bar{a}_t$ have different values for each element in RAD as shown in (\ref{eq:mu}). This makes a difference in the generation process where $\mu_{\theta,t}$ is utilized.} In fact, RAD solves a more difficult problem than that of DDPM, because $\epsilon_t$ becomes more complicated. Hence, using this exact same structure might not be as successful as in DDPM, as also confirmed in our ablation study. 

To resolve this issue, we focus on the $t$ input in $\epsilon_\theta(x_t,t)$, which helps the network handle different steps adaptively. In practice, $t$ undergoes some embedding, comprising a cos-sin encoding and FC layers, and is added to every pixel of feature maps in the U-Net. Regarding this $t$ input, we conjecture that its main role is to inform $\epsilon_\theta$ of the overall intensity of $\epsilon_t$. This is somewhat reasonable because the intensity of $\epsilon_t$ increases as $t$ progresses in DDPM. Accordingly, we propose to use $\bar{b}_t$, the pixel-wise intensity of $\epsilon_t$ in RAD, in place of $t$ instead. This is easily accomplished by replacing the FC layers in the embedding module into $1\times1$ convolutions. In this way, the pixel-wise noise condition can be directly informed to the network, making it easier to learn the spatially variant $\epsilon_t$. This actually works without changing any other component, confirming the above conjecture.

The above spatial noise embedding can be viewed as an indirect way of spatial conditioning, which has some similarities with conventional conditioning techniques. However, it is more subtle in that it alters the existing $t$ embedding and is completely determined by the noise schedules.

\subsection{Practical Considerations} \label{sec:practical}

Although RAD has been explained in terms of DDPM, the proposed approach based on spatially variant noise is not limited to DDPM; it can be similarly applied to other advanced models such as DDIM \cite{song2020denoising}, iDDPM \cite{nichol2021improved}, ADM \cite{dhariwal2021diffusion}, score-based models \cite{song2021scorebased}, and stable diffusion \cite{rombach2022high}. 
We have successfully tested RAD with many of these. For example, DDIM shares the same training procedure as DDPM, but differs in the reverse process, which we can also apply a similar element-wise reformulation.
Similarly, iDDPM and ADM share the forward and reverse processes of DDPM but differ in the denoiser structures and loss functions, which combine (\ref{eq:reduced_VLB}) and (\ref{eq:simplified_loss}). Hence, RAD can be extended to these without an issue, and all examples in the experiments are based on ADM versions. Since the basic framework of RAD is quite simple and universal, we expect it to work on other similar models as well.

One limitation of RAD is that the alternate formulation requires fresh training. Even though this is the reason for the superb generation speed compared to the existing SoTA methods, it can also be a deal-breaker considering the amount of resources required in training. This is why the most recent methods attempt to modify only the generation processes of pretrained models, sacrificing the speed. In this paper, we address this issue by utilizing LoRA \cite{hu2021lora}, which can significantly reduce training efforts by leveraging pretrained models. To train RAD based on LoRA, some minor adjustment is required:
The spatial noise embedding in RAD can be too drastic of a change for LoRA.
Fortunately, the noise schedules for individual pixels are set to match that in DDPM, so we can inverse-map the elements of $\bar{b}_t$ to timestep values based on the accumulated noise levels in the steps of DDPM. We use linear interpolation in this process, meaning that the mapped timestep values can be non-integral.
This has proven to be effective in the experiments, where all the examples are actually fine-tuned by LoRA based on pretrained ADMs. This approach makes RAD much more accessible.

In an actual implementation of RAD, singularities in the loss function must be carefully reviewed and avoided. In fact, there is a potential singularity in the variational loss (\ref{eq:reduced_VLB}), especially in the first step ($t=1$) where $p_\theta(x_0 | x_1)$ is prone to becoming degenerate. Even though DDPM uses the simplified loss (\ref{eq:simplified_loss}) in practice, which has no singularities, later works such as iDDPM and ADM use both (\ref{eq:reduced_VLB}) and (\ref{eq:simplified_loss}), where the singularities become a problem and are handled by some heuristics. This is an often-overlooked problem that can render training unstable, and for RAD, the issue is a little more complicated because the pixels undergo asynchronous noise schedules. The details are explained in the supplementary material.

\section{Experiments}
\label{sec:experiments}

\subsection{Benchmark Datasets}
We validate RAD on the FFHQ \cite{karras2019style}, LSUN Bedroom \cite{yu2015lsun}, and ImageNet \cite{deng2009imagenet}. These datasets are well-suited for training and evaluating high-quality image generation models.
\begin{itemize}
    \item \textbf{FFHQ (Flickr-Faces-HQ)} \cite{karras2019style} contains a total of 70,000 high-resolution (1024x1024) images of human faces. This dataset provides greater diversity than traditional facial image datasets, encompassing a wide range of ages, genders, ethnicities, hairstyles, and accessories (e.g., glasses, hats, etc.). All images were resized to $256 \times 256$. 
    
    \item \textbf{LSUN Bedroom} \cite{yu2015lsun} is a large-scale dataset of indoor scenes, containing over 3M bedroom images. Captured from various angles and perspectives, these images represent complex room structures, enabling algorithms to learn and generate intricate scene layouts. We train RAD with 288K samples of the LSUN Bedroom for convenience. All images were resized to $256 \times 256$.

    \item \textbf{ImageNet} \cite{deng2009imagenet} is a large-scale image dataset for visual object recognition, containing over 1.2M labeled images across 1,000 categories. This dataset offers extensive diversity, covering a wide range of objects, animals, scenes, and complex visual compositions. All images were resized to $256 \times 256$  for consistency in our experiments.

\end{itemize}

\begin{table*}[!t]
    \footnotesize
    \centering
    \caption{Performance (FID and LPIPS) on FFHQ and LSUN Bedroom with various mask types. $^\dagger$ indicates that the value is quoted from the original paper. \textbf{Bold}: best, \underline{under}: second best.}
    \begin{tabular}{lcccccccccccc}
        \toprule
        \multirow{3}{*}{Method}& \multicolumn{6}{c}{FFHQ} & \multicolumn{6}{c}{LSUN Bedroom} \\
        \cmidrule(lr){2-7} \cmidrule(lr){8-13}
        & \multicolumn{2}{c}{Box} & \multicolumn{2}{c}{Extreme} & \multicolumn{2}{c}{Wide} & \multicolumn{2}{c}{Box} & \multicolumn{2}{c}{Extreme} & \multicolumn{2}{c}{Wide} \\
        & FID $\downarrow$ & LPIPS $\downarrow$ & FID $\downarrow$ & LPIPS $\downarrow$ & FID $\downarrow$ & LPIPS $\downarrow$ & FID $\downarrow$ & LPIPS $\downarrow$ & FID $\downarrow$ & LPIPS $\downarrow$ & FID $\downarrow$ & LPIPS $\downarrow$ \\
        \midrule
        LaMa \cite{suvorov2022resolution} & 27.7$^\dagger$ & \underline{0.086}$^\dagger$ & 61.7$^\dagger$ & 0.492$^\dagger$ & 23.2$^\dagger$ &\underline{0.096}$^\dagger$ & - & - & - & - & - & - \\
        Score-SDE \cite{song2021scorebased} & 30.3$^\dagger$ & 0.135$^\dagger$ & 48.6$^\dagger$ & 0.488$^\dagger$ & 29.8$^\dagger$ & 0.132$^\dagger$ & 23.7 & 0.648 & 24.1 & 0.648 & 23.2 & 0.644 \\
        DDRM \cite{kawar2022denoising} & 28.4$^\dagger$ & 0.109$^\dagger$ & 48.1$^\dagger$ & 0.532$^\dagger$ & 27.5$^\dagger$ & 0.113$^\dagger$ & 20.5 & 0.166 & 33.1 & 0.450 & 26.4 & 0.190 \\
        RePAINT \cite{lugmayr2022repaint} & 25.7$^\dagger$ & 0.093$^\dagger$ & 35.9$^\dagger$ & 0.398$^\dagger$ & 24.2$^\dagger$ & 0.108$^\dagger$ & 20.5 & 0.176 & 23.5 & 0.461 & 21.4 & 0.161 \\
        MCG \cite{chung2022improving}& \underline{23.7}$^\dagger$ & 0.089$^\dagger$ & \textbf{30.6}$^\dagger$ & 0.366$^\dagger$ & \underline{22.1}$^\dagger$ & 0.099$^\dagger$ & \underline{19.9} & \underline{0.131} & \underline{22.0} & \textbf{0.395} & \underline{20.9} & \underline{0.108}\\
        \midrule
        DDNM \cite{wang2023ddnm} & 30.4 & 0.089 & 87.7 & \underline{0.353} & 30.4 & 0.089 & 22.7 & 0.150 & 53.3 & 0.431 & 23.2 & 0.126 \\
        DeqIR \cite{cao2024deep} & 24.2 & 0.093 & 64.2 & 0.368 & 27.4 & 0.099 & 22.2 & 0.176 & 43.9 & 0.461 & 22.0 & 0.153 \\
        \midrule
        RAD (ours) & \textbf{22.1} & \textbf{0.074} & \underline{33.4} & \textbf{0.317} & \textbf{21.5} & \textbf{0.078} & \textbf{19.2} & \textbf{0.131} & \textbf{21.6} & \underline{0.399} & \textbf{20.8} & \textbf{0.107}\\
        \bottomrule
    \end{tabular}
    \label{tab:comparison}
\end{table*}

\begin{figure*}[!t]
    \scriptsize
    \centering
    \addtolength{\tabcolsep}{-6pt}
    \begin{subfigure}{0.32\linewidth}
        \begin{tabular}{cccc}
            \textbf{Input} & \textbf{RePaint} & \textbf{MCG} & \textbf{RAD}\\
            \includegraphics[width=0.25\linewidth]{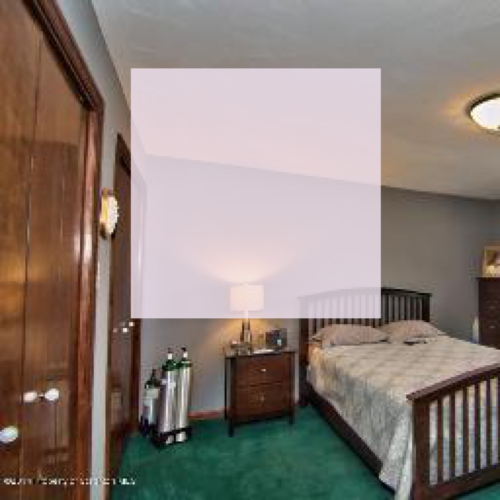} &
            \includegraphics[width=0.25\linewidth]{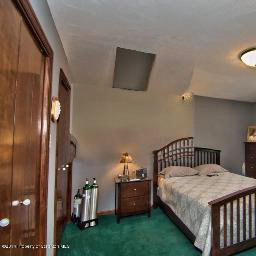} &
            \includegraphics[width=0.25\linewidth]{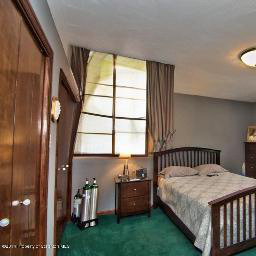} &
            \includegraphics[width=0.25\linewidth]{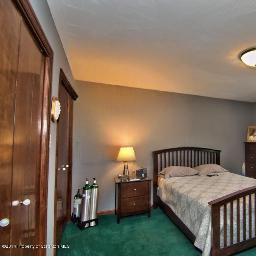} \\
            \includegraphics[width=0.25\linewidth]{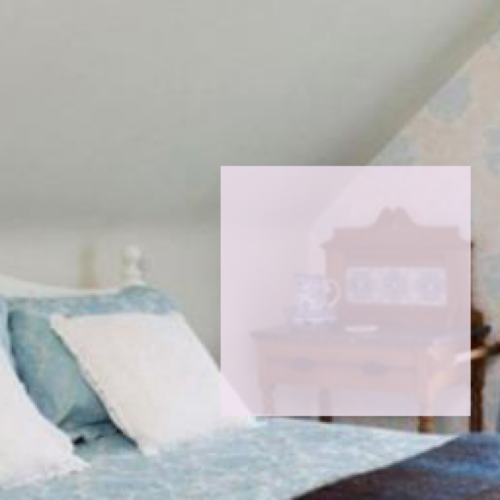} &
            \includegraphics[width=0.25\linewidth]{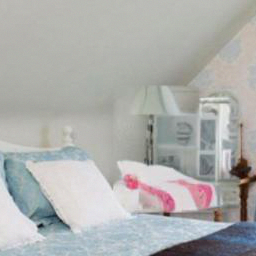} &
            \includegraphics[width=0.25\linewidth]{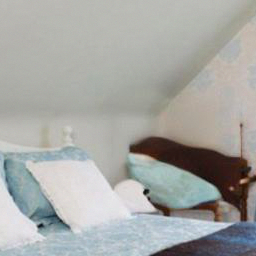} &
            \includegraphics[width=0.25\linewidth]{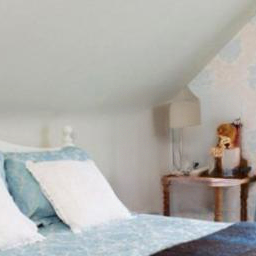} \\
            \includegraphics[width=0.25\linewidth]{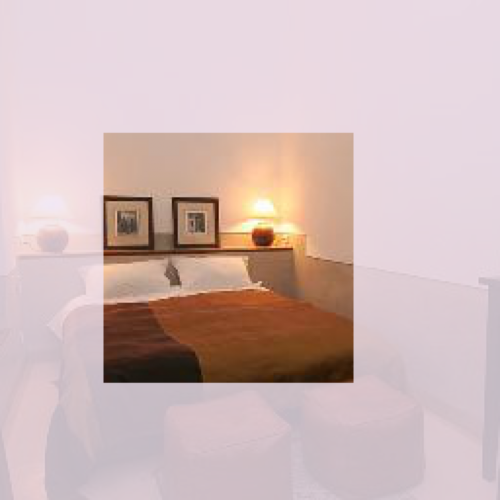} &
            \includegraphics[width=0.25\linewidth]{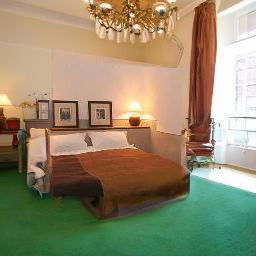} &
            \includegraphics[width=0.25\linewidth]{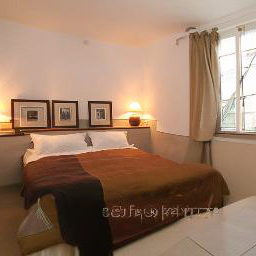} &
            \includegraphics[width=0.25\linewidth]{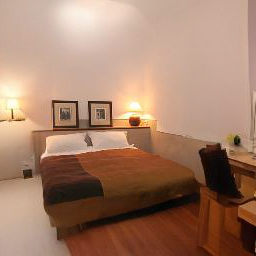} \\
            \includegraphics[width=0.25\linewidth]{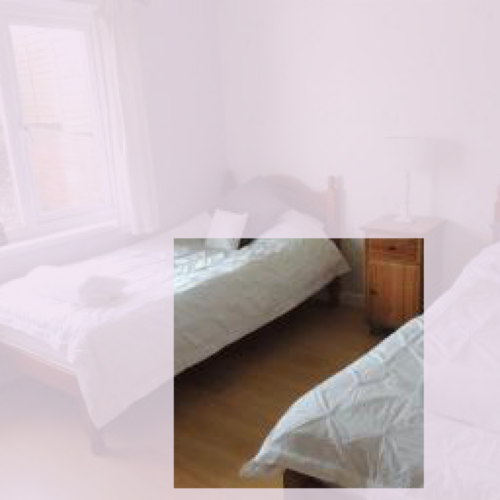} &
            \includegraphics[width=0.25\linewidth]{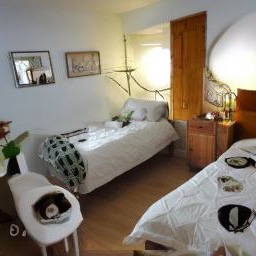} &
            \includegraphics[width=0.25\linewidth]{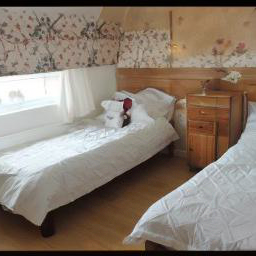} &
            \includegraphics[width=0.25\linewidth]{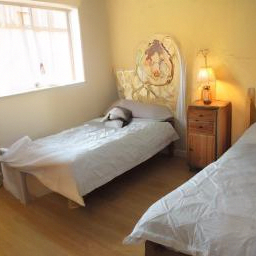} \\
            \includegraphics[width=0.25\linewidth]{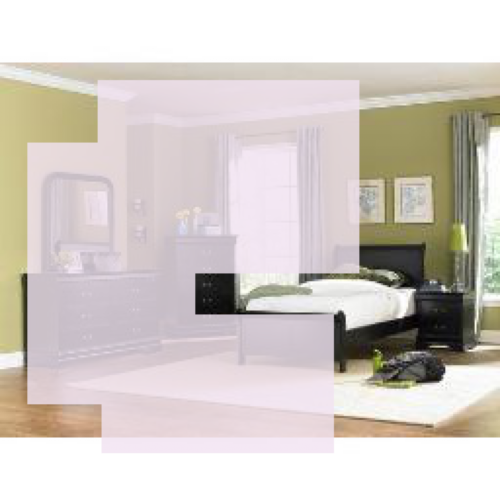} &
            \includegraphics[width=0.25\linewidth]{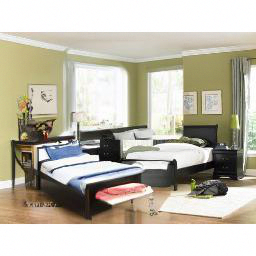} &
            \includegraphics[width=0.25\linewidth]{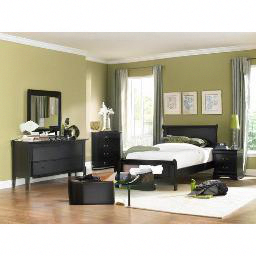} &
            \includegraphics[width=0.25\linewidth]{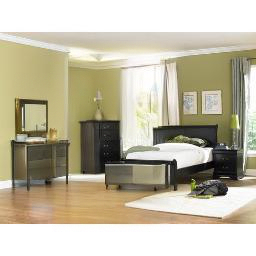} \\
            \includegraphics[width=0.25\linewidth]{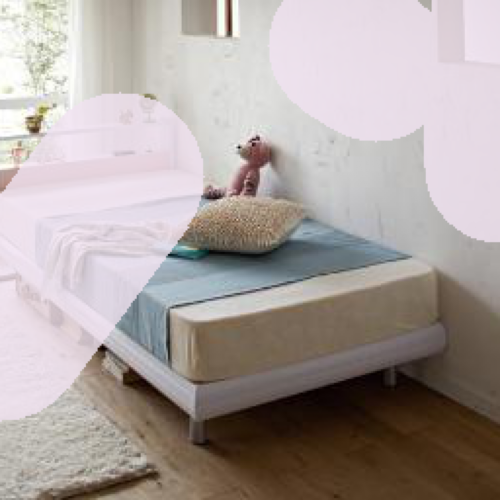} &
            \includegraphics[width=0.25\linewidth]{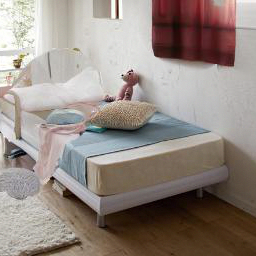} &
            \includegraphics[width=0.25\linewidth]{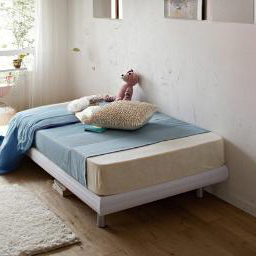} &
            \includegraphics[width=0.25\linewidth]{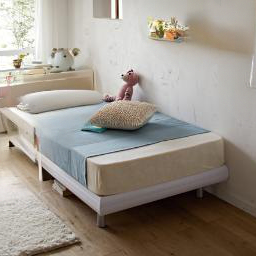}
        \end{tabular}\caption{LSUN Bedroom} \label{fig:comparison:Bedroom}
    \end{subfigure}
    \hfill
    \begin{subfigure}{0.32\linewidth}
        \begin{tabular}{cccc}
            \textbf{Input} & \textbf{RePaint} & \textbf{MCG} & \textbf{RAD}\\
            \includegraphics[width=0.25\linewidth]{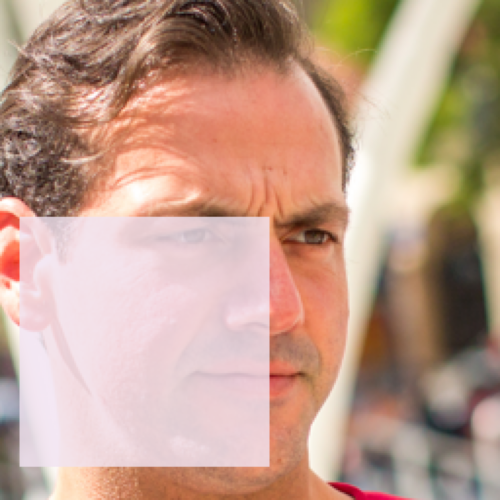} &
            \includegraphics[width=0.25\linewidth]{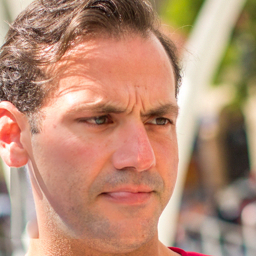} &
            \includegraphics[width=0.25\linewidth]{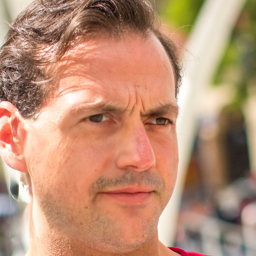} &
            \includegraphics[width=0.25\linewidth]{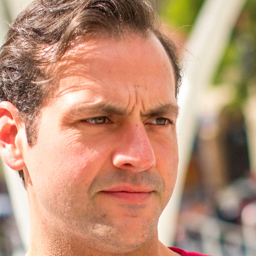} \\
            \includegraphics[width=0.25\linewidth]{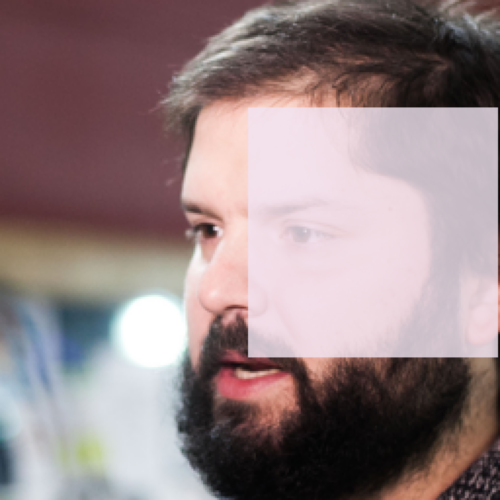} &
            \includegraphics[width=0.25\linewidth]{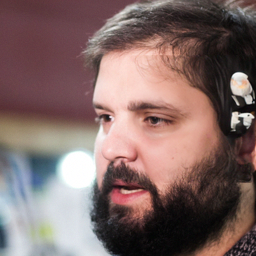} &
            \includegraphics[width=0.25\linewidth]{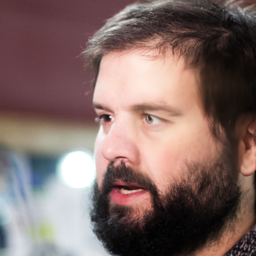} &
            \includegraphics[width=0.25\linewidth]{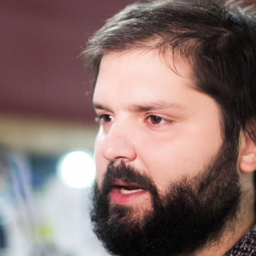} \\
            \includegraphics[width=0.25\linewidth]{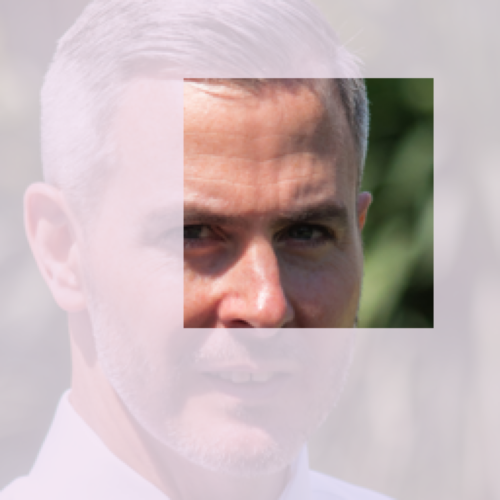} &
            \includegraphics[width=0.25\linewidth]{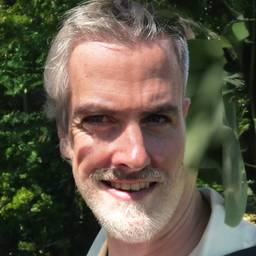} &
            \includegraphics[width=0.25\linewidth]{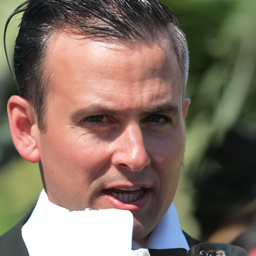} &
            \includegraphics[width=0.25\linewidth]{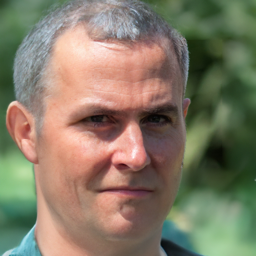} \\
            \includegraphics[width=0.25\linewidth]{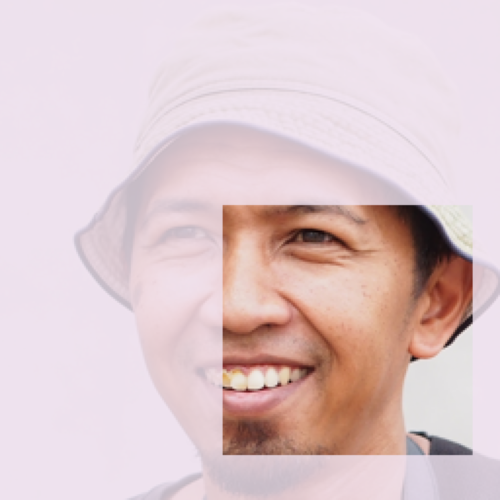} &
            \includegraphics[width=0.25\linewidth]{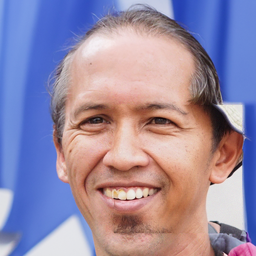} &
            \includegraphics[width=0.25\linewidth]{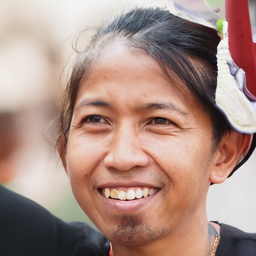} &
            \includegraphics[width=0.25\linewidth]{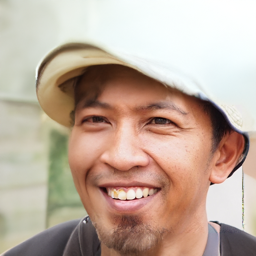} \\
            \includegraphics[width=0.25\linewidth]{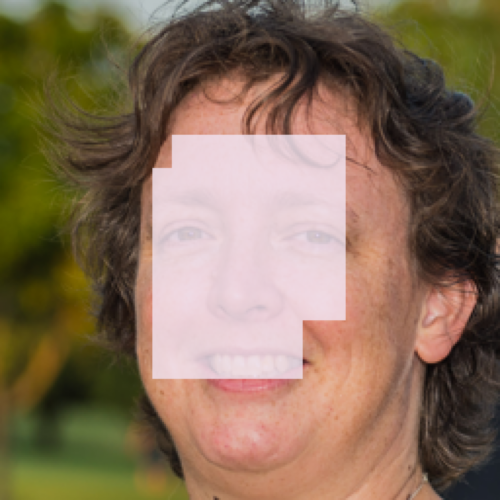} &
            \includegraphics[width=0.25\linewidth]{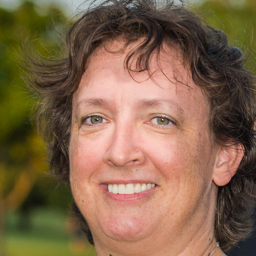} &
            \includegraphics[width=0.25\linewidth]{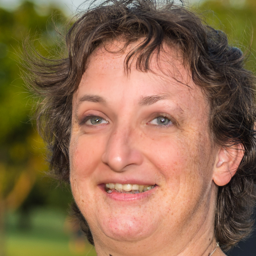} &
            \includegraphics[width=0.25\linewidth]{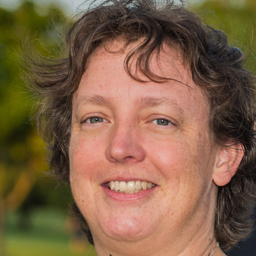} \\
            \includegraphics[width=0.25\linewidth]{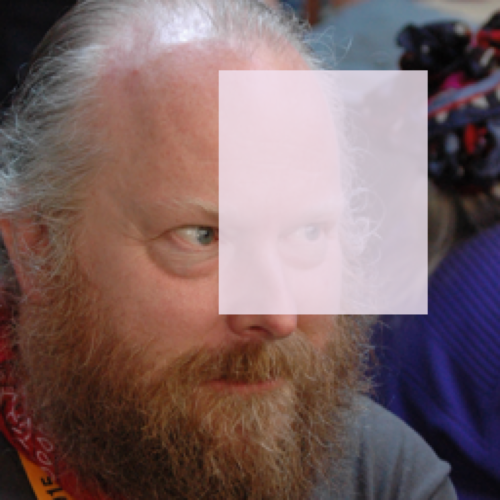} &
            \includegraphics[width=0.25\linewidth]{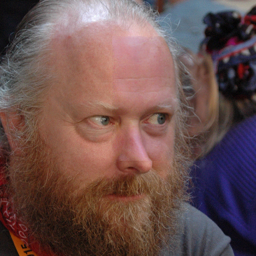} &
            \includegraphics[width=0.25\linewidth]{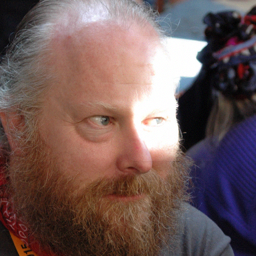} &
            \includegraphics[width=0.25\linewidth]{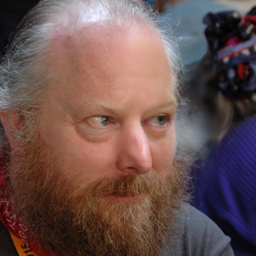} \\
        \end{tabular} \caption{FFHQ}\label{fig:comparison:FFHQ}
    \end{subfigure}
    \hfill
    \begin{subfigure}{0.32\linewidth}
        \begin{tabular}{cccc}
            \textbf{Input} & \textbf{RePaint} & \textbf{MCG} & \textbf{RAD}\\
            \includegraphics[width=0.25\linewidth]{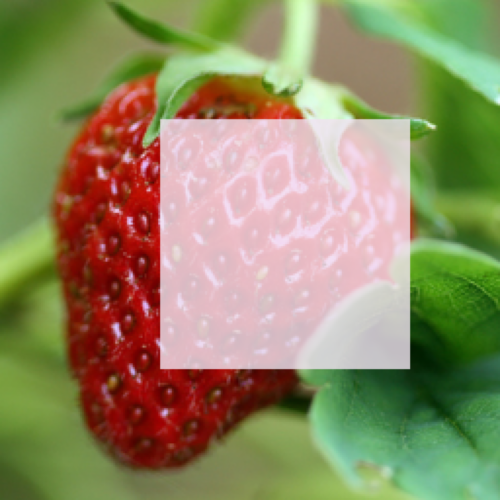}  &
            \includegraphics[width=0.25\linewidth]{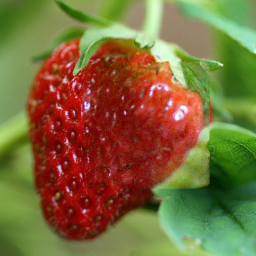} &
            \includegraphics[width=0.25\linewidth]{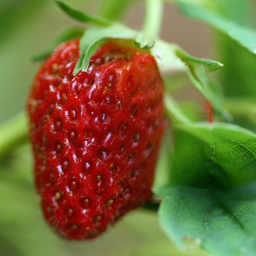} &
            \includegraphics[width=0.25\linewidth]{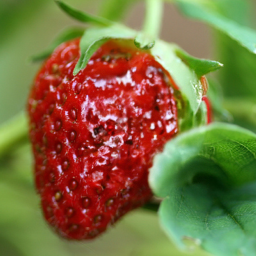} \\
            \includegraphics[width=0.25\linewidth]{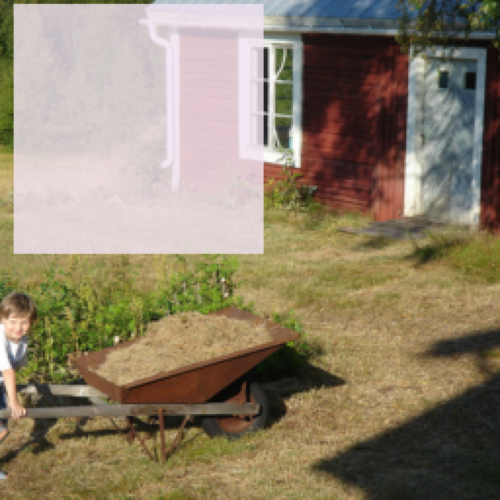}  &
            \includegraphics[width=0.25\linewidth]{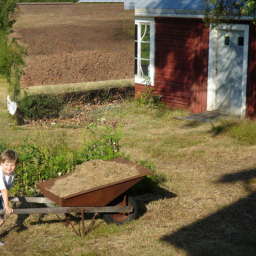} &
            \includegraphics[width=0.25\linewidth]{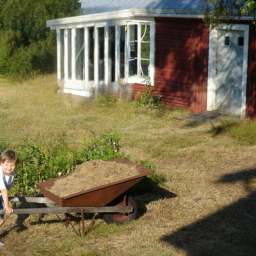} &
            \includegraphics[width=0.25\linewidth]{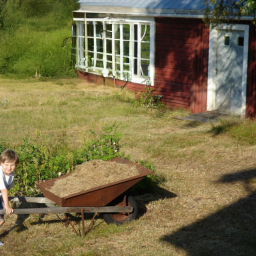} \\
            \includegraphics[width=0.25\linewidth]{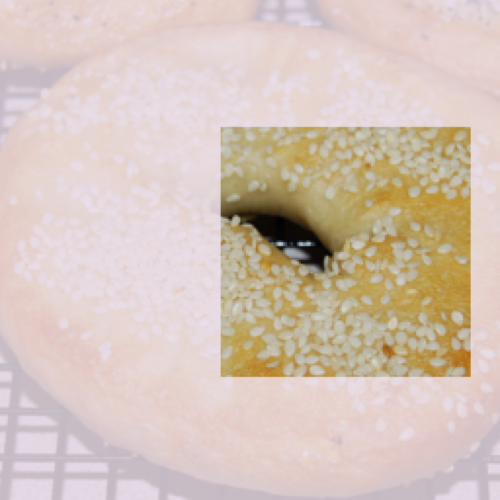}  &
            \includegraphics[width=0.25\linewidth]{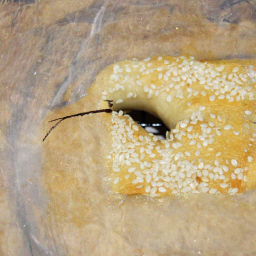} &
            \includegraphics[width=0.25\linewidth]{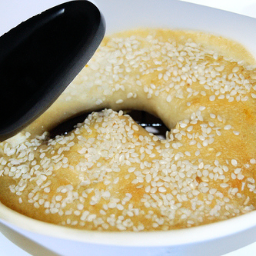} &
            \includegraphics[width=0.25\linewidth]{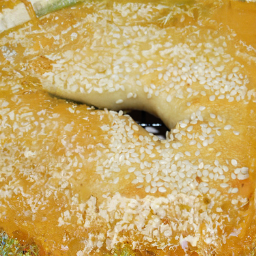} \\
            \includegraphics[width=0.25\linewidth]{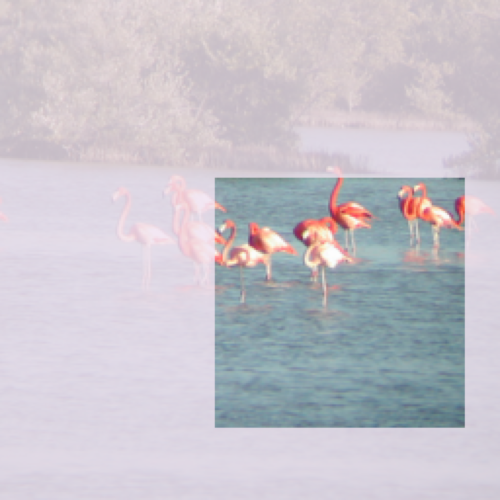}  &
            \includegraphics[width=0.25\linewidth]{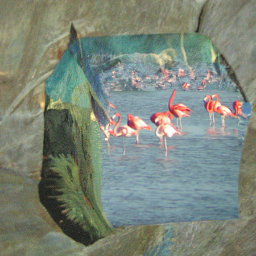} &
            \includegraphics[width=0.25\linewidth]{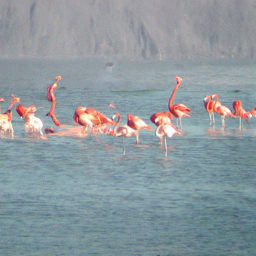} &
            \includegraphics[width=0.25\linewidth]{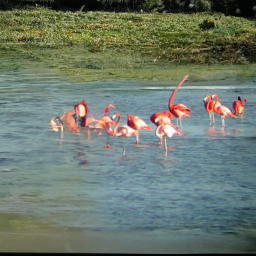} \\
            \includegraphics[width=0.25\linewidth]{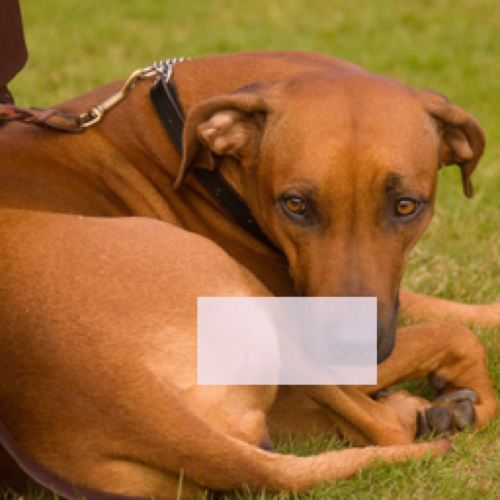}  &
            \includegraphics[width=0.25\linewidth]{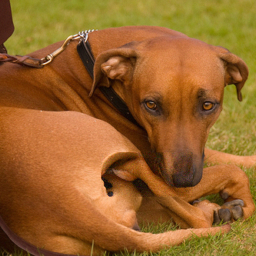} &
            \includegraphics[width=0.25\linewidth]{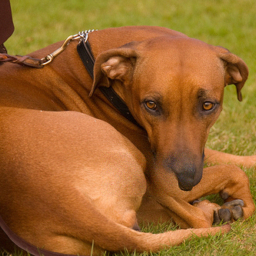} &
            \includegraphics[width=0.25\linewidth]{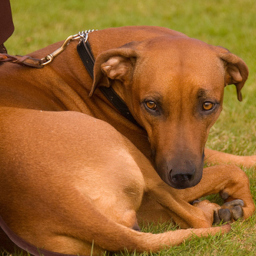} \\
            \includegraphics[width=0.25\linewidth]{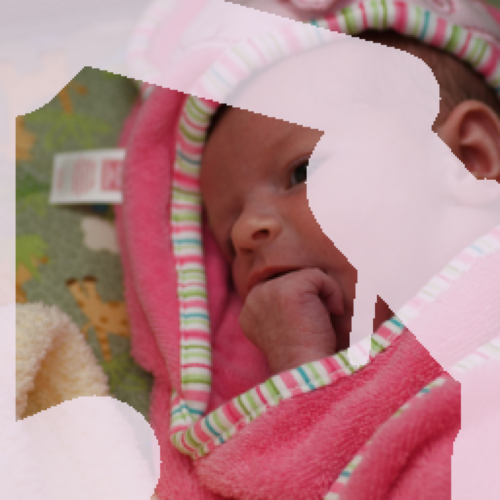}  &
            \includegraphics[width=0.25\linewidth]{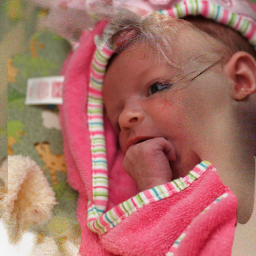} &
            \includegraphics[width=0.25\linewidth]{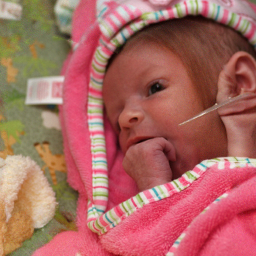} &
            \includegraphics[width=0.25\linewidth]{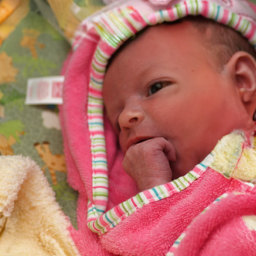} \\
        \end{tabular} \caption{ImageNet}\label{fig:comparison:ImageNet}
    \end{subfigure}
        \caption{Qualitative comparisons. Colored areas indicate inpainting regions (1st/2nd rows: box, 3rd/4th: extreme, 5th/6th: wide).}
    \label{fig:comparison}
    \vspace{-3mm}
\end{figure*}

\subsection{Implementation Details}
\paragraph{Training settings.}
For all experiments, we set the rank of LoRA \cite{hu2021lora} to 16 and trained RAD using the Adam optimizer \cite{DBLP:journals/corr/KingmaB14} with a learning rate of $10^{-4}$. We used a batch size of 16 for both FFHQ and ImageNet, while that was eight for LSUN Bedroom. The total diffusion steps was 2000, with the initial 1000 steps dedicated to Phase 1. We used predefined linear noise schedules for $\beta_t, \beta_t' \in \left(0.0001, 0.02\right)$. During inpainting, we used 100 sampling steps. FFHQ was trained on four TITAN RTX GPUs, LSUN Bedroom on eight NVIDIA 3090 GPUs, and ImageNet on eight NVIDIA RTX 6000 ADA GPUs. Pretrained ADM models \cite{dhariwal2021diffusion} were used for all datasets. RAD was trained for 300K on FFHQ and LSUN Bedroom, and 800K on ImageNet.

\vspace{-4mm}
\paragraph{Validation settings.}
For validation, we set aside 1000 images from all datasets.
We evaluated RAD across three types of mask configurations: box, extreme, and wide, following the practice in \cite{suvorov2022resolution}. A box mask randomly removes a $128 \times 128$ square region. In contrast, an extreme mask retains only a specific $128 \times 128$ region, removing all other surrounding areas. A wide mask defines irregular missing regions. For inpainting, we used the reverse process of DDPM rather than DDIM due to its superior quality in generating realistic details.

\vspace{-4mm}
\paragraph{Baseline.}
We chose various inpainting methods based on GANs and diffusion models. For the GAN-based baseline, we used LaMA \cite{suvorov2022resolution}, and for the diffusion-based baselines, we employed RePaint \cite{lugmayr2022repaint}, MCG \cite{chung2022improving}, DDRM \cite{kawar2022denoising}, DDNM \cite{wang2023ddnm}, and DeqIR \cite{cao2024deep}. Publicly available pretrained ADM models \cite{dhariwal2021diffusion} were used for all methods. Additionally, each method was evaluated using its publicly available codebase to maintain consistency in performance assessment.

\vspace{-4mm}
\paragraph{Metrics.}
To evaluate the performance of the proposed method and compare it with existing approaches, we employed Frechet Inception Distance (FID) \cite{heusel2017gans} and the Learned Perceptual Image Patch Similarity (LPIPS) \cite{zhao2020uctgan} as the inpainting quality measure.


\subsection{Results}
\paragraph{Comparisons.}
\Cref{tab:comparison} shows the inpainting performance of various methods. 
Reported values are either directly quoted from the original papers or computed by us if publicly available pretrained models exist.
For LaMa, no pretrained model was available for LSUN Bedroom.
Here, RAD achieves the best performance in terms of LPIPS and demonstrates superior FID scores for FFHQ and LSUN Bedroom compared to most baseline methods. Notably, DDNM and DeqIR exhibit relatively weaker performance in our experiments, even though they are relatively newer methods. The main goal of these methods is image restoration, and they are not specifically designed for challenging inpainting scenarios, particularly those involving large missing areas. Consequently, these methods show worse performance under our experimental settings. \Cref{fig:comparison} shows
qualitative comparisons on LSUN Bedroom, FFHQ, and ImageNet. Here, all the methods are compared based on the same combinations of images and inpainting masks. The results confirm that RAD generally produces more natural images than others and has fewer failure cases.

\Cref{tab:runtime} shows the inference time of various methods. We evaluated all methods on $256 \times 256$ images using a single NVIDIA TITAN RTX GPU.
Here, RAD is about 100 times faster than RePaint and 15 times faster than MCG while delivering superior performance. 
Baseline methods other than RePaint and MCG are much faster, but their inpainting quality is significantly worse.

\begin{table}[!t]
    \footnotesize
    \centering
    \caption{Inference time comparison ($1\times$ NVIDIA TITAN RTX).}
    \begin{tabular}{lc}
        \toprule
        Method & Inference time [s] \\
        \midrule
        Score-SDE \cite{song2021scorebased} & 41.983 \\
        DDRM \cite{kawar2022denoising} & 12.291 \\
        RePAINT \cite{lugmayr2022repaint} & 837.136 \\
        MCG \cite{chung2022improving} & 128.058 \\
        \midrule
        DDNM \cite{wang2023ddnm} & 6.566 \\
        DeqIR \cite{cao2024deep} & 26.443 \\
        \midrule
        Ours & \textbf{8.442} \\
        \bottomrule
    \end{tabular}
    \label{tab:runtime}
\end{table}

\begin{figure}[!t]
    \centering
    \footnotesize
    \addtolength{\tabcolsep}{-6pt}
    \begin{tabular}{cccccc}
        \textbf{Input} & \textbf{Sample 1} & \textbf{Sample 2} & \textbf{Sample 3} & \textbf{Sample 4}\\

        \includegraphics[width=0.08\textwidth]{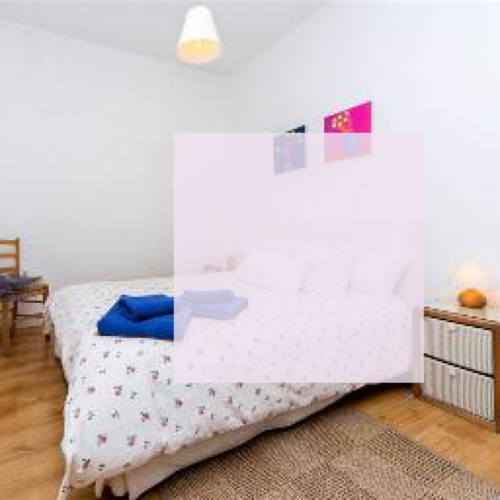}  & \includegraphics[width=0.08\textwidth]{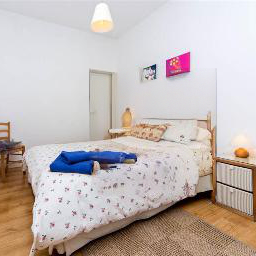} & \includegraphics[width=0.08\textwidth]{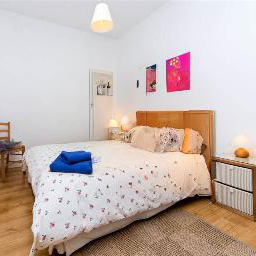} & \includegraphics[width=0.08\textwidth]{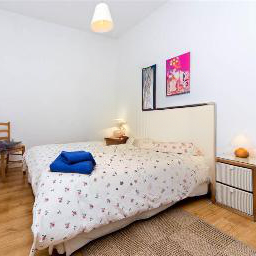} &\includegraphics[width=0.08\textwidth]{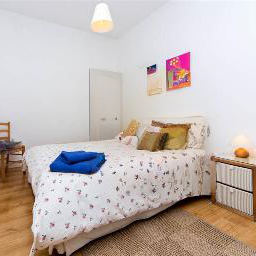} \\
        \includegraphics[width=0.08\textwidth]{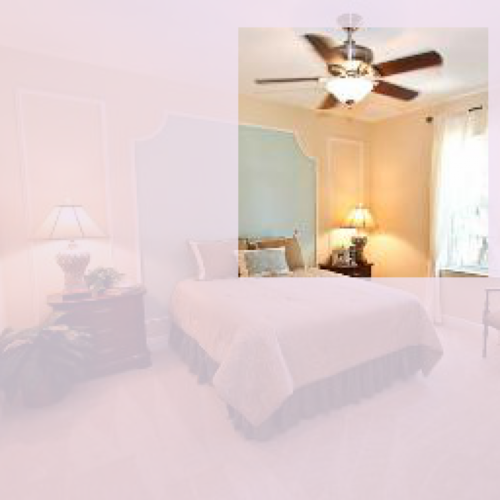} & \includegraphics[width=0.08\textwidth]{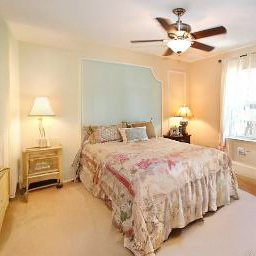} & \includegraphics[width=0.08\textwidth]{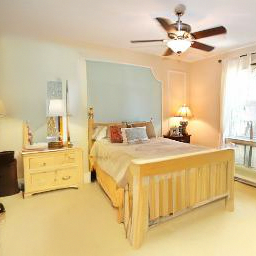} & \includegraphics[width=0.08\textwidth]{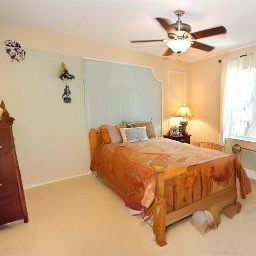} &\includegraphics[width=0.08\textwidth]{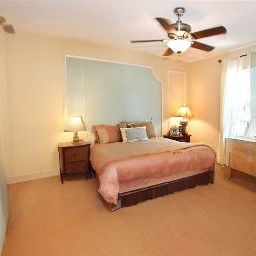}  \\
        \includegraphics[width=0.08\textwidth]{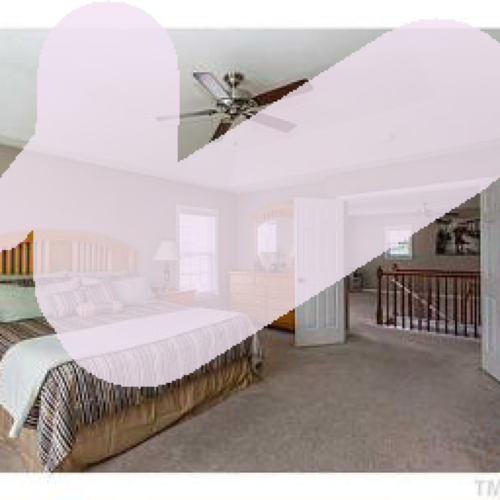}   & \includegraphics[width=0.08\textwidth]{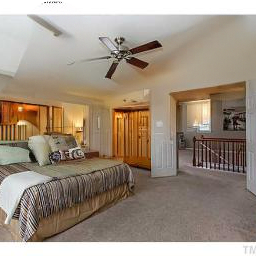} & \includegraphics[width=0.08\textwidth]{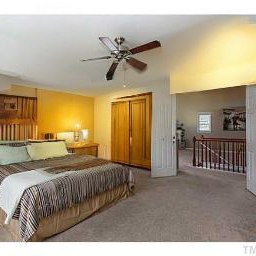} & \includegraphics[width=0.08\textwidth]{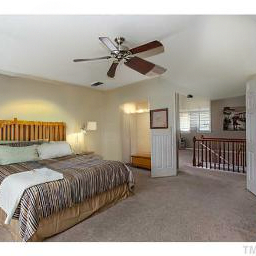}&\includegraphics[width=0.08\textwidth]{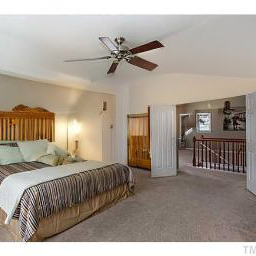}  \\
        \includegraphics[width=0.08\textwidth]{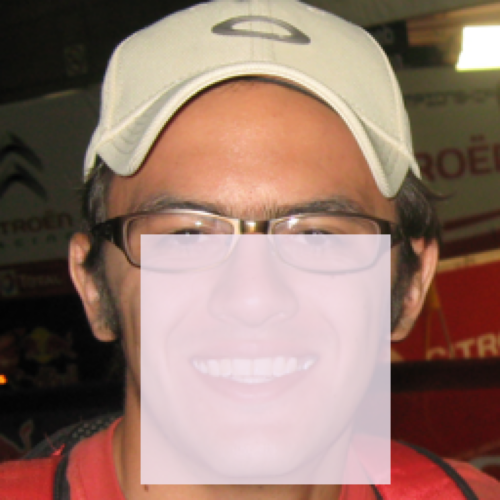} &\includegraphics[width=0.08\textwidth]{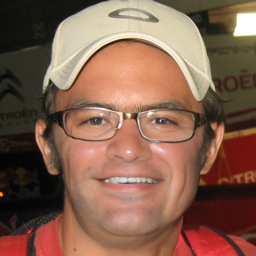} & \includegraphics[width=0.08\textwidth]{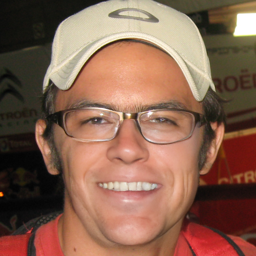} & \includegraphics[width=0.08\textwidth]{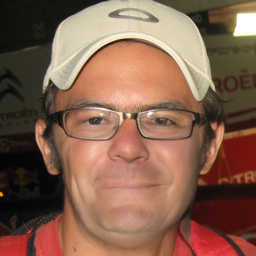}&\includegraphics[width=0.08\textwidth]{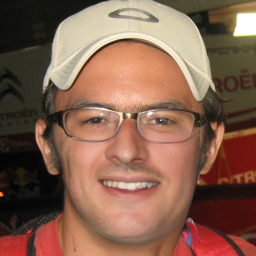} \\
        \includegraphics[width=0.08\textwidth]{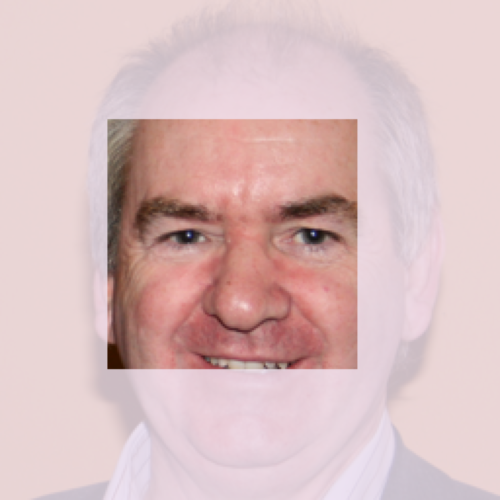} &\includegraphics[width=0.08\textwidth]{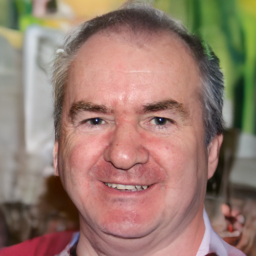} & \includegraphics[width=0.08\textwidth]{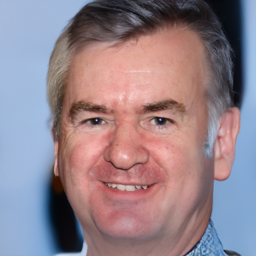} & \includegraphics[width=0.08\textwidth]{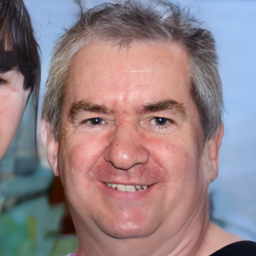}&\includegraphics[width=0.08\textwidth]{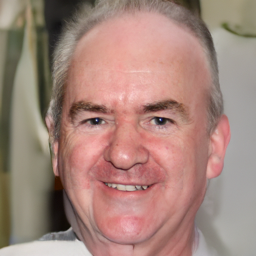}  \\
        \includegraphics[width=0.08\textwidth]{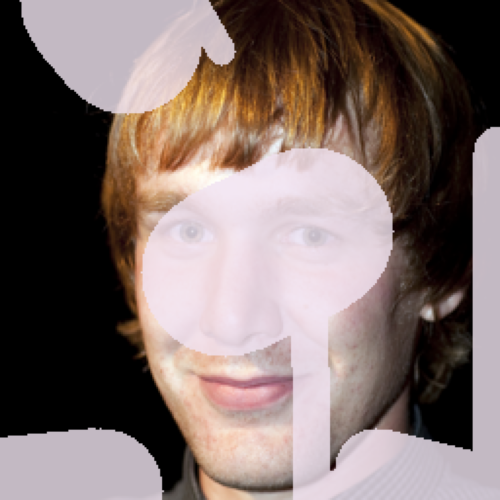} &\includegraphics[width=0.08\textwidth]{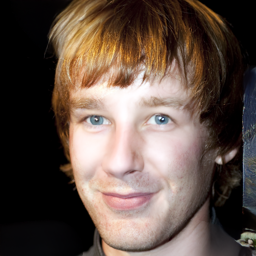} & \includegraphics[width=0.08\textwidth]{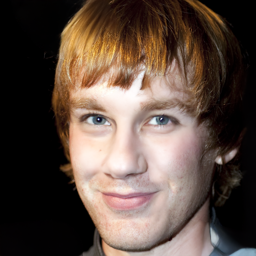} & \includegraphics[width=0.08\textwidth]{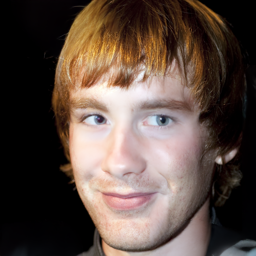}&\includegraphics[width=0.08\textwidth]{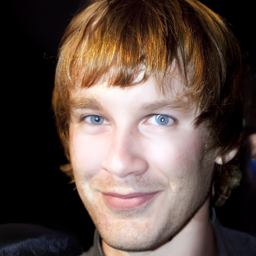}   \\
        
    \end{tabular}
        \caption{Example results of RAD on LSUN Bedroom and FFHQ.}
    \label{tab:multiple_samples}
\end{figure}

\vspace{-4mm}
\paragraph{Analysis of diversity.}
\Cref{tab:multiple_samples} demonstrates that RAD can generate diverse inpaintings across various mask configurations. Each row in the figure shows the input image with a specific mask, followed by multiple inpainting results (Samples 1 to 4) generated by RAD. These results highlight RAD's ability to produce various plausible and contextually consistent inpaintings, even under challenging masks.

\begin{table}[!t]
    \small
    \centering
    \caption{Ablation study of RAD on FFHQ (Cfg. 1: pretrained ADM with RAD reverse steps, Cfg. 2: w/o spatial noise emb.).}
    \resizebox{\linewidth}{!}{
    \begin{tabular}{p{1.5cm}cccccccc}
    \toprule
        \multirow{2}{*}{Method} & \multicolumn{2}{c}{Box} & \multicolumn{2}{c}{Extreme} & \multicolumn{2}{c}{Wide}  \\
        & FID $\downarrow$ & LPIPS $\downarrow$ & FID $\downarrow$ & LPIPS $\downarrow$ & FID $\downarrow$ & LPIPS $\downarrow$ \\
        \midrule
        Cfg. 1  & 128.3 & 0.279 & 172.0 & 0.441 & 95.7 & 0.240\\
        Cfg. 2  & 23.5 & 0.086 & 37.0 & 0.333 & 26.8 & 0.085\\
        \midrule
        RAD (ours) & \textbf{22.1} & \textbf{0.074} & \textbf{33.4} & \textbf{0.317} & \textbf{21.5} & \textbf{0.078} \\
        \bottomrule
    \end{tabular}
    }
    \label{tab:ablation quantitative}
    \vspace{-2mm}
\end{table}

\begin{figure}[!t]
    \centering
    \scriptsize
    \addtolength{\tabcolsep}{-6pt}
    \begin{tabular}{ccccccccc}
        \textbf{Input}  & \textbf{Cfg. 1} & \textbf{Cfg. 2} & \textbf{RAD} \\
        \includegraphics[width=0.09\textwidth]{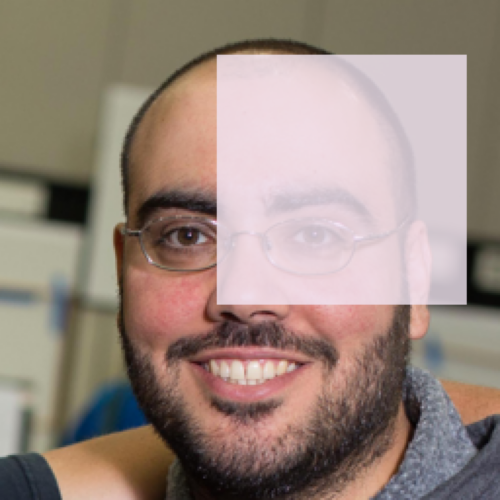} &\includegraphics[width=0.09\textwidth]{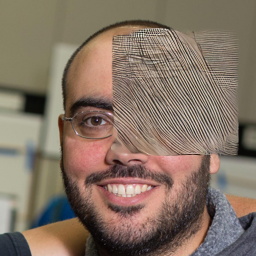} & \includegraphics[width=0.09\textwidth]{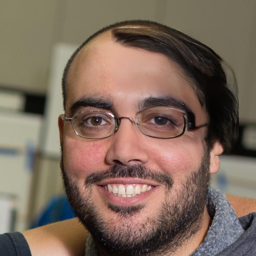} & \includegraphics[width=0.09\textwidth]{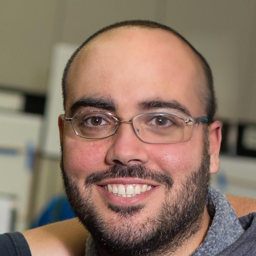} \\
        \includegraphics[width=0.09\textwidth]{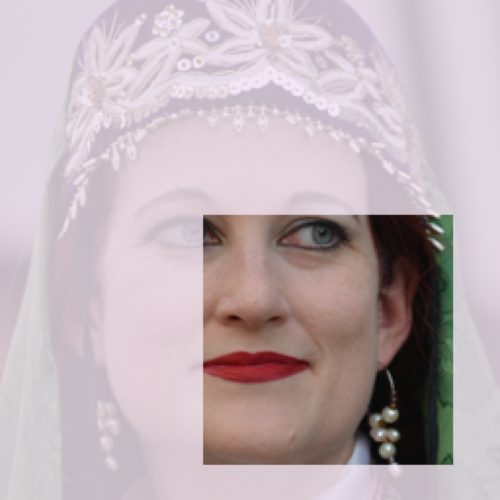} &\includegraphics[width=0.09\textwidth]{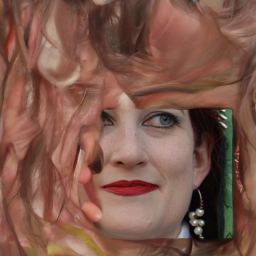} & \includegraphics[width=0.09\textwidth]{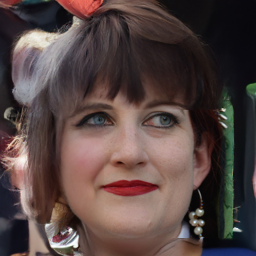} & \includegraphics[width=0.09\textwidth]{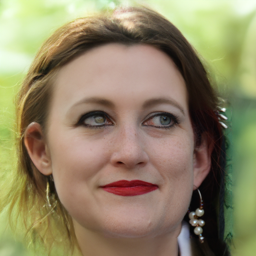} \\ 
        \includegraphics[width=0.09\textwidth]{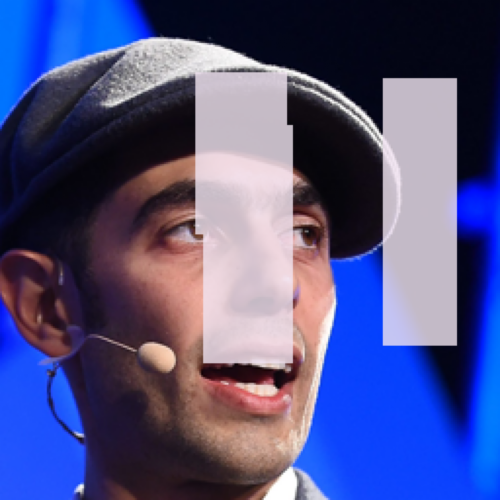} &\includegraphics[width=0.09\textwidth]{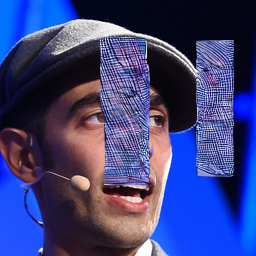} & \includegraphics[width=0.09\textwidth]{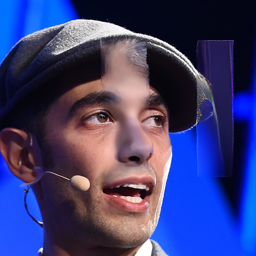} & \includegraphics[width=0.09\textwidth]{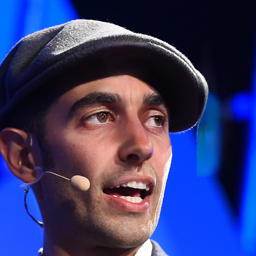} 
        
    \end{tabular}
        \caption{Ablation study examples on FFHQ (Cfg. 1: pretrained ADM with RAD reverse steps, Cfg. 2: w/o spatial noise emb.).}
    \label{fig:ablation_qualitative}
    \vspace{-1mm}
\end{figure}

\vspace{-4mm}
\paragraph{Ablation study.}
To verify the effectiveness of the proposed components, we compared RAD with two alternative methods: (1) performing RAD reverse steps directly on a pretrained ADM (\ie, no spatially variant noise training and no spatial noise embedding) and (2) RAD without spatial noise embedding (\ie, using $t$ embedding).
As shown in \Cref{tab:ablation quantitative}, the RAD reverse steps do not perform well on a pretrained ADM model, which suggests that training with spatially variant noise is vital for RAD. Moreover, $t$ embedding exhibits inferior performance to spatial noise embedding in RAD, showing the effectiveness of the proposed embedding in inpainting tasks. This conclusion is further supported by qualitative results in \Cref{fig:ablation_qualitative}.

\section{Limitations}
A limitation of RAD is that it requires explicit training. The burden of training a diffusion model can be significant, and accordingly, we utilized LoRA to mitigate this issue greatly.
Additionally, RAD is trained using a spatially variant noise schedule defined based on masks, making it dependent on the mask distribution. Perlin masks proposed in this paper are effective enough to handle most inpainting scenarios. However, there is a possibility that RePaint or MCG might yield better results in drastic cases even though they take much longer inference time.

\section{Conclusion}
\label{sec:conclusion}
We presented a novel inpainting model, RAD, which utilizes a spatially variant noise schedule to allow asynchronous generation of different regions. RAD handles inpainting by selectively adding noise to the regions specified by a given mask, followed by denoising those areas. This approach enables the selective generation of masked regions while preserving the others. 
The Perlin noise-based mask generation technique has been presented to generate realistic masks during training. Additionally, we proposed spatial noise embedding, significantly enhancing inpainting quality compared to conventional $t$ embedding. RAD can perform seamless inpainting without any additional module, and a plain reverse process is sufficient to produce high-quality results with orders of magnitude faster sampling speed.

While RAD is mainly described in the context of inpainting, the overall framework has broader implications.
By extending the pseudo-targets of diffusion models (the denoising distributions) to spatially variant distributions, RAD offers more means to analyze and manipulate spatial interactions in diffusion models, opening interesting future research directions. Combining RAD with conditions such as text will extend the framework to image editing, a direction for future work.
{
    \small
    \bibliographystyle{ieeenat_fullname}
    \bibliography{main}
}


\end{document}